\documentclass{article}

\PassOptionsToPackage{numbers, compress}{natbib}
\usepackage[final]{neurips_data_2024}
\usepackage[toc,page,header]{appendix}
\usepackage{minitoc}

\usepackage[utf8]{inputenc} 
\usepackage[T1]{fontenc}    
\usepackage{hyperref}       
\usepackage{url}            
\usepackage{booktabs}       
\usepackage{amsfonts}       
\usepackage{nicefrac}       
\usepackage{microtype}      
\usepackage{multirow}
\usepackage{array}
\usepackage{boldline}
\usepackage[table, x11names]{xcolor}
\usepackage{epsfig}

\setcitestyle{square,comma}
\hypersetup{
  colorlinks,
  citecolor=black,
  linkcolor=black,
  urlcolor=black}

\usepackage{float}
\usepackage{listings}
\usepackage{enumitem}
\usepackage{wrapfig}

\title{NorwAI's Large Language Models: Technical Report}

\author{%
  Jon Atle Gulla\thanks{All authors are listed in alphabetical order by last name. \textbf{Contributors:} Terje Brasethvik, Kerstin Bach, Simen Eide, Anders Haarr, Jon Espen Ingvaldsen, Benjamin Kille.} \quad Peng Liu$^{*}$ \quad Lemei Zhang$^{*}$ \\ Norwegian Research Center for AI Innovation (NorwAI), NTNU\\ \texttt{\{jon.atle.gulla, peng.liu, lemei.zhang\}@ntnu.no} \\ \url{https://huggingface.co/NorwAI}
}

\begin{document}

\maketitle

\begin{abstract}
Norwegian, spoken by approximately five million people, remains underrepresented in many of the most significant breakthroughs in Natural Language Processing (NLP). To address this gap, the NorLLM team at NorwAI has developed a family of models specifically tailored to Norwegian and other Scandinavian languages, building on diverse Transformer-based architectures such as GPT, Mistral, Llama2, Mixtral and Magistral. These models are either pretrained from scratch or continually pretrained on ~25B $-$ 88.45B tokens, using a Norwegian-extended tokenizer and advanced post-training strategies to optimize performance, enhance robustness, and improve adaptability across various real-world tasks. Notably, instruction-tuned variants (e.g., Mistral-7B-Instruct and Mixtral-8x7B-Instruct) showcase strong assistant-style capabilities, underscoring their potential for practical deployment in interactive and domain-specific applications. The NorwAI large language models are openly available to Nordic organizations, companies and students for both research and experimental use. This report provides detailed documentation of the model architectures, training data, tokenizer design, fine-tuning strategies, deployment, and evaluations.
\end{abstract}

\section{Introduction}
Large Language Models (LLMs) have become foundational technologies in Natural Language Processing (NLP), demonstrating remarkable capabilities in tasks such as text generation \cite{brown2020language,raffel2020exploring}
, summarization \cite{zhang2024benchmarking,zhang2024personalsum}, translation \cite{he2024exploring}, and reasoning \cite{guo2025deepseek}. Despite these advances, the development and support of LLMs for low-resource languages like Norwegian remain limited \cite{kummervold2021operationalizing,kutuzov2021large}. Norway's linguistic landscape includes Bokmål, Nynorsk, and Sámi, alongside a pervasive English influence. In smaller linguistic communities, language technology is often dominated by English-trained systems. Such imbalance can lead to biases, loss of linguistic nuance, and diminished cultural representation \cite{samuel2023norbench,de2025impact}. Developing localized LLMs is therefore essential for preserving linguistic diversity, enabling domain-specific applications, and ensuring national digital sovereignty.

NorwAI, Norway's national research center for artificial intelligence, has spearheaded efforts to address this challenge. Leveraging high-performance computing resources at NTNU Idun, NorwAI has trained a suite of competitive Norwegian LLMs, adapting open architectures like Mistral and Llama to Norwegian data. By developing Norwegian LLMs, NorwAI ensures accessibility, transparency, and relevance of AI for both public and private sectors in Norway.

These models support a wide range of practical applications, including education, media, healthcare, and government services. This report provides a detailed technical account, serving both as documentation and as a baseline for future research in Norwegian language modeling.

In summary, our contributions can be summarized as follows:
\begin{itemize}
  \item We have developed the largest and most diverse suite of fundamental Norwegian generative language models\footnote{\url{https://github.com/Smartmedia-AI/NorGLM}} \cite{liu2024nlebench} to date, trained on the NTNU HPC cluster Idun. These models cover a range of architectures (GPT, Llama, Mistral, Mixtral, Magistral) and parameter sizes (350M, 3B, 7B, 23B, 45B).
  \item We construct a comprehensive Norwegian pre-training corpus, mainly focusing on publicly available textual data as well as high-quality news and transcription data from our industry partners, such as Schibsted and NRK.
  \item We introduce a new benchmark dataset, NLEBench \cite{liu2024nlebench}, for evaluating generative language modeling in Norwegian. NLEBench encompasses a broad spectrum of real-world NLP tasks, ranging from news storytelling, summarization, open-domain conversation, and natural language understanding to human-annotated instruction fine-tuning, toxicity and bias evaluation, and multi-task learning.
\end{itemize}

\section{Available Models}
All NorwAI LLMs are hosted on Hugging Face under the NorwAI (\url{https://huggingface.co/NorwAI}
) and NorGLM (\url{https://huggingface.co/NorGLM}
) organizations and can be downloaded free of charge for research and experimental use. Each model repository provides pretrained model weights, configuration files, tokenizer vocabularies, and quantized versions to facilitate efficient inference and experimentation. Detailed usage instructions, quantization formats, and hosting options are described in Section~\ref{sec:deployment_inference}, \emph{Deployment Inference}.

The current NorwAI model family spans multiple parameter scales and architectures, providing a flexible foundation for Norwegian and broader Scandinavian language processing. The available models and their release dates are as follows:
\begin{itemize}
\item \textbf{NorGPT-369M (Initial release: 2023):} A 369M-parameter Transformer trained from scratch, serving as a lightweight baseline.
\item \textbf{NorGPT-3B (Initial release: 2023):} A compact 3B-parameter Transformer model, trained from scratch and optimized for efficiency and accessibility.
\item \textbf{NorGPT-23B (Initial release: 2023):} A large 23B-parameter Transformer trained from scratch, designed for document-level comprehension.
\item \textbf{NorLlama-3B (Initial release: 2023):} A 3B-parameter Llama-style model trained from scratch for Norwegian.
\item \textbf{NorwAI-Mistral-7B-base/instruct (Initial release: 2024):} A 7.5B-parameter model built on the Mistral architecture with sliding-window attention, trained in both base (from-scratch/continual pretraining) and instruction-tuned variants.
\item \textbf{NorwAI-Llama2-7B (Initial release: 2024):} A Norwegian adaptation of Meta's Llama2-7B, continually pretrained on curated Norwegian corpora.
\item \textbf{NorwAI-Mixtral-8x7B-base/instruct (Initial release: 2024):} A Mixture-of-Experts (MoE) model comprising 8 experts (7B parameters each, 47B in total), with 2 experts activated per token. It is available in both base and instruction-tuned variants, designed for efficient inference and long-document understanding.
\item \textbf{NorwAI-Magistral-24B (NorwAI's first reasoning model. Initial release: 2025):} A 23.9B-parameter Transformer with a 158k-token vocabulary and 40k context, continually pretrained on Mistral's Magistral, that adaptively switches between Completion, Short Thinking, and Long Thinking modes for balanced efficiency and performance while avoiding unnecessary overthinking.
\end{itemize}

\noindent\textbf{Running the models.}
The models can be run directly with common open-source frameworks such as Hugging Face Transformers, vLLM, or Ollama. A minimal workflow employs Hugging Face Transformers with $device\_map=``auto"$ to distribute layers across available devices, as illustrated in Section~\ref{sec:deployment_inference_hf}. Quantized GGUF builds (Q3-Q8) are also supported, reducing memory requirements at a small cost in precision. Beyond the standard Transformers interface, models can be deployed using high-throughput servers such as \emph{vLLM}, local or lightweight API hosting via \emph{Ollama}, or integrated into managed infrastructures such as the \emph{Telenor AI Factory} and \emph{SINTEF LM Studio}. Typical (order-of-magnitude) GPU memory requirements are as follows: the smallest NorGPT-369M model runs comfortably on a 4~GB GPU or a modern CPU; models around 3~billion parameters typically require 12-16~GB of GPU memory; 7B-class models (Mistral and Llama2) perform best on 24-32~GB GPUs such as the A100 or RTX~4090; the Magistral-24B reasoning model requires approximately 64-80~GB; and the Mixture-of-Experts model Mixtral-8x7B is designed for distributed multi-GPU setups with around 115~GB or more total memory.

\noindent\textbf{Conditions for use.}
Consistent with the project's goal of open access for the Nordic region, the models are available to organizations, companies, and students in the Nordic countries for research and experimental use. Users should always consult the license terms on each model card for specific permissions and attribution requirements.

\noindent\textbf{Disclaimer.}
The NorwAI LLMs are provided \emph{as is}, without warranties of any kind. No model has been specifically aligned to prevent the generation of toxic or otherwise offensive content, and outputs may in some cases be inappropriate or harmful. Use is at the user's own risk. Users are responsible for ensuring legal and ethical compliance, as well as for validating model outputs before integrating them into any application.

\section{Model Training}
\subsection{Model Configurations}
NorwAI LLMs employ Transformer-based architectures with several enhancements to improve efficiency and performance on Norwegian text. The model suite includes decoder-only Transformer models (e.g. GPT, Llama2, Mistral) and sparsely-activated expert models (e.g. Mixture-of-Experts, MoE). Each model adapts existing architectures with customized tokenizer vocabularies (64k, 68k, 158k tokens) and parameter settings (350M, 3B, 7B, 23B, 45B). Large models handle contexts up to 32k tokens, ideal for document-level comprehension, while smaller models like NorGPT-3B maintain shorter contexts for latency-sensitive applications. Sparse activation in MoE models enables scaling parameter counts without proportionally increasing computation. Furthermore, memory-optimized attention mechanisms (e.g., FlashAttention) and fused kernels enhance training efficiency on large GPU clusters.

Table~\ref{tab:models} summarizes their specifications, with each column highlighting a key characteristic of the NorwAI language models. \textbf{Params} indicates the number of trainable parameters, reflecting model capacity and computational cost. \textbf{Architecture} specifies the neural design, such as Transformer or Mixture-of-Experts (MoE), which determines efficiency, scalability, and activation sparsity. \textbf{Context Length} shows the maximum number of tokens each model can process at once; longer contexts enable document-level comprehension and long-form reasoning. \textbf{Hidden Size} refers to the dimensionality of token vectors, which determines the model's representational capacity and per-layer computation. \textbf{Tokenizer Vocab size} refers to the number of subword units in the tokenizer, influencing linguistic coverage and efficiency for Norwegian text. \textbf{Training Strategy} describes whether a model was trained from scratch, continually pretrained, or instruction-tuned, clarifying its degree of specialization and target applications. The \textbf{GPU Hours} column reports the total compute time used during model training, providing an estimate of computational cost and resource scale. \textbf{Accelerator} specifies the GPU hardware employed, such as A100 or H100, which affects training speed, memory bandwidth, and energy efficiency. Together, these columns provide a comprehensive overview of the NorwAI model suite's scale, architecture, computational resources, and training methodology.

\begin{table}
\footnotesize
  \caption{NorwAI LLMs training specifications.}
  \label{tab:models}
  \centering
  \setlength\tabcolsep{2.6pt}
  \begin{tabular}{p{3.96cm}p{1.0cm}p{1.6cm}<{\centering}p{1.0cm}<{\centering}p{0.8cm}<{\centering}p{0.8cm}<{\centering}p{2.72cm}p{0.72cm}<{\raggedleft}p{1.4cm}<{\centering}}
    \toprule
     \multirow{2}*{Model} & \multirow{2}*{Params} & \multirow{2}*{Architecture} & Context Length & Hidden Size & Vocab Size & \multirow{2}*{Training Strategy} & GPU Hours & Accelerator (NVIDIA) \\ 
    \hlineB{1.5}
    \href{https://huggingface.co/NorGLM/NorGPT-369M}{NorGPT-369M} & 369.6M & Transformer & 2k & 1024 & 64k & Trained from scratch & 1.3K & A100\\
    \href{https://huggingface.co/NorGLM/NorGPT-3B}{NorGPT-3B} & 2.95B & Transformer & 2k & 2688 & 64k &  Trained from scratch & 2.6K & A100  \\
    NorGPT-23B  & 23.03B & Transformer & 2k & 6144 & 64k &  Trained from scratch & 53.1K & A100\\
    \href{https://huggingface.co/NorGLM/NorLlama-3B}{NorLlama-3B} & 3.07B & Transformer & 2k & 2688 & 64k &  Trained from scratch & 2.2K & A100 \\
    \href{https://huggingface.co/NorwAI/NorwAI-Mistral-7B-pretrain}{NorwAI-Mistral-7B-pretrain} & 7.54B & Transformer & 4k & 4096 & 68k &  Trained from scratch & 5.1K & H100 \\
    \href{https://huggingface.co/NorwAI/NorwAI-Mistral-7B}{NorwAI-Mistral-7B} & 7.54B & Transformer & 4k & 4096 & 68k &  Continual Pretraining & 4.6K & H100\\
    \href{https://huggingface.co/NorwAI/NorwAI-Llama2-7B}{NorwAI-Llama2-7B} & 7.03B & Transformer & 4k & 4096 & 68k &  Continual Pretraining & 4.4K & H100 \\
    \href{https://huggingface.co/NorwAI/NorwAI-Mixtral-8x7B}{NorwAI-Mixtral-8x7B} & 47.00B & MoE & 32k & 4096 & 68k &  Continual Pretraining & 31.1K & H100 \\
    \href{https://huggingface.co/NorwAI/NorwAI-Magistral-24B-reasoning}{NorwAI-Magistral-24B} & 23.85B & Transformer & 40k & 5120 & 158k &  Continual Pretraining & 22K & H100\\
    \href{https://huggingface.co/NorwAI/NorwAI-Mistral-7B-instruct}{NorwAI-Mistral-7B-instruct} & 7.54B & Transformer & 4k & 4096 & 68k & Instruction tuning & 38.4 & H100\\
    \href{https://huggingface.co/NorwAI/NorwAI-Mixtral-8x7B-instruct}{NorwAI-Mixtral-8x7B-instruct} & 47.00B & MoE & 32k & 4096 & 68k & Instruction tuning & 170.8 & H100\\
    \bottomrule
  \end{tabular}
  \vspace{-6pt}
\end{table}

\subsection{Pre-training}

Pretraining relies on a carefully curated Norwegian corpus. For our initial corpus (\emph{NorLLM\_Corpus\_V1}) created in 2023, we filtered Norwegian texts from the mC4 and OSCAR web-crawled corpora and included non-copyrighted material from the National Library of Norway (Nasjonalbiblioteket) \cite{kummervold2021operationalizing}. We also sourced high-quality news articles from Schibsted and collected tweets (January 2012 to December 2022) as well as Reddit posts (October 2017 to December 2022) via their respective APIs. To enhance robustness in downstream tasks, we incorporated Danish, Swedish, and German texts, along with a small portion of English, all sourced from the mC4 corpus. The total corpus size was $25$B tokens.

In 2024, we removed the social media data and added more high-quality news articles from Schibsted, VG, and NRK, bringing the corpus (\emph{NorLLM\_Corpus\_V2}) to $51.15$B tokens. In our latest corpus (\emph{NorLLM\_Corpus\_V3}), compiled early in 2025, we replaced the Norwegian material from Nasjonalbiblioteket with their updated Mímir core, which includes the public portion of the Norwegian Colossal Corpus (NCC) with permissible licenses, Bokmål and Nynorsk CulturaX, and Bokmål and Nynorsk HPLT v$1.2$. We further enriched the corpus with high-quality transcription data from NRK, additional Sámi text resources, and domain-specific data, increasing the total size to $88.45$B tokens. An overview of the data sources and their corresponding trained models is shown in Table~\ref{tab:corpora_models}.

Over successive iterations, the share of copyrighted material in our Norwegian corpus has declined substantially, while the volume of industry data from our partners has increased, as shown in Figure 1A. Moreover, the amount of Nynorsk and Sámi text continues to expand, rising from $1.124$\% to $5.217$\% for Nynorsk and from $0$\% to $0.155$\% for Sámi in the Norwegian portion, as shown in Figure 1B.

Preprocessing includes deduplication, language identification, normalization, and sentence segmentation. Tokenizers are trained using SentencePiece with vocabulary sizes scaling from 64k to 158k tokens, allowing efficient representation of Norwegian morphology and dialectal variation.

We utilized LLM Foundry\footnote{\url{https://github.com/mosaicml/llm-foundry/tree/main}} as a codebase for training and fine-tuning NorwAI LLMs, except for NorGPT and NorLlama-3B models. LLM Foundry, developed by MosaicML, incorporates the Composer\footnote{\url{https://github.com/mosaicml/composer}} library to implement distributed training workflows on large-scale clusters. NorGPT models were trained with the Megatron-DeepSpeed Framework\footnote{\url{https://github.com/deepspeedai/Megatron-DeepSpeed}}, while NorLlama-3B employed the Tencent Pre-training Framework\footnote{\url{https://github.com/Tencent/TencentPretrain}}.

\begin{figure}
\centering
\epsfig{figure=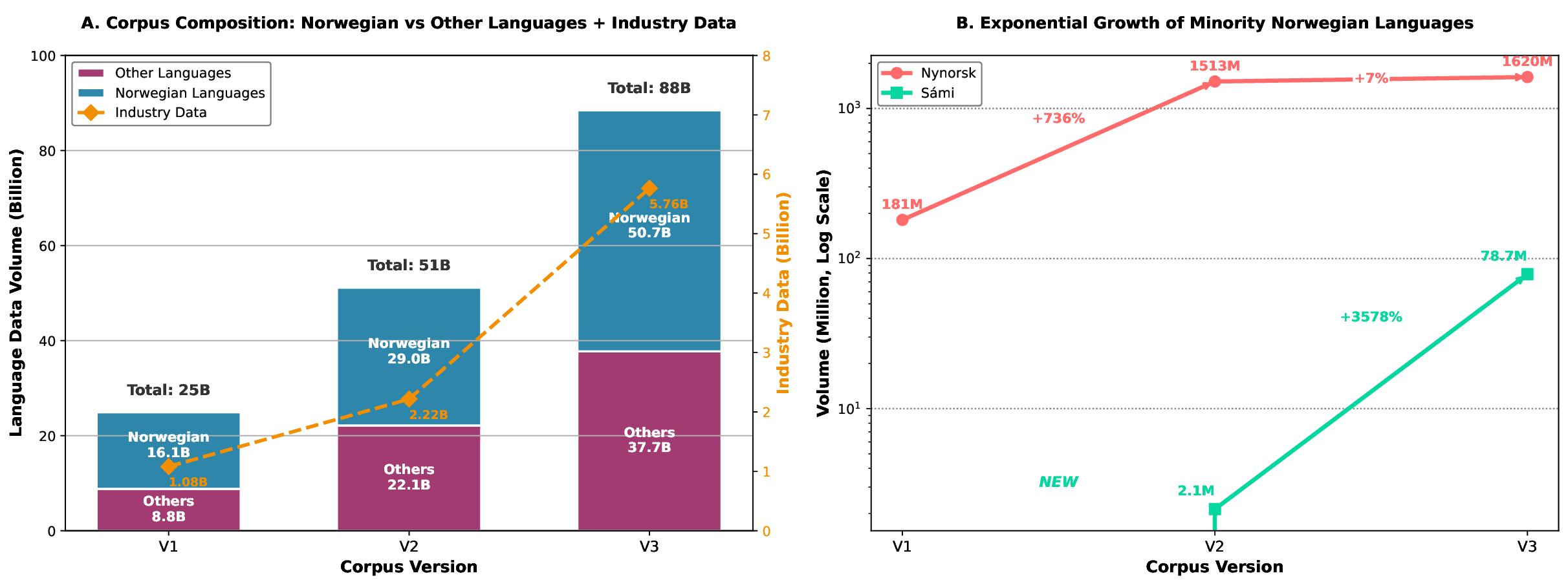, width=437pt, height=183pt}
\caption{Statistics of the NorLLM training corpora, with the number of tokens measured in billions (A) and millions (B).} \label{fig:supplementary_bench}
\end{figure}

\begin{table}
\footnotesize
  \caption{Overview of NorLLM corpus data sources and corresponding trained models (in billions of tokens). Industry partners include Schibsted, VG, and NRK. Common Crawl sources comprise mC4, OSCAR, CulturaX, and HPLT.}
  \label{tab:corpora_models}
  \centering
  \setlength\tabcolsep{3.2pt}
  \begin{tabular}{p{3.1cm}p{2.4cm}<{\centering}p{1.7cm}<{\centering}p{1.3cm}<{\centering}p{1.0cm}<{\centering}p{0.2cm}p{4.3cm}}
    \toprule
     \multirow{3}*{Corpora} & \multicolumn{4}{c}{Data Sources} & & \multirow{3}*{Associated Model(s)} \\ 
     \cmidrule[0.8pt]{2-5} & \multirow{2}*{Nasjonalbiblioteket} & Industry Partners & Common Crawl & \multirow{2}*{Others} & & \\ 
    \cmidrule[0.8pt]{1-7}
    \multirow{2}*{\emph{NorLLM\_Corpus\_V1}} & \multirow{2}*{3.38B} & \multirow{2}*{1.08B} & \multirow{2}*{20.41B} & \multirow{2}*{0.13B} & & NorGPT-369M/3B/23B, NorLlama-3B \\ \cmidrule[0.6pt]{1-7}
    \multirow{3}*{\emph{NorLLM\_Corpus\_V2}} & \multirow{3}*{9.45B} & \multirow{3}*{2.22B} & \multirow{3}*{36.00B} & \multirow{3}*{3.33B} &  & NorwAI-Mistral-7B-base/instruct, NorwAI-Llama2-7B, NorwAI-Mixtral-8x7B-base/instruct \\ \cmidrule[0.6pt]{1-7}
    \emph{NorLLM\_Corpus\_V3}  & 21.88B & 5.76B & 51.78B & 8.58B & & NorwAI-Magistral-24B \\
    \bottomrule
  \end{tabular}
  \vspace{-5pt}
\end{table}

\subsection{Post-training}

\subsubsection{Instruction Tuning and Alignment}

Instruction tuning is a crucial step in aligning LLMs with human objectives, ensuring that they follow user instructions accurately and generate contextually appropriate responses. For Norwegian LLMs, this process combines high-quality annotated instruction data with parameter-efficient fine-tuning methods such as LoRA \cite{hulora}, resulting in models that are robust, efficient, and well adapted to the Norwegian language, culture, and diverse applications.

\noindent\textbf{Fine-Tuning on High-Quality Instruction Data} We first fine-tuned the full model weights on a curated instruction dataset\footnote{\url{https://huggingface.co/datasets/mimir-project/mimir-instruction}}, which contains nearly 5000 instructions annotated by our research assistants. The dataset spans various aspects of Norwegian culture, daily life, history, and geography. Tasks within the dataset include summarization, question answering, single-choice and multiple-choice questions, and other formats. Representative examples from this dataset are provided in the Appendix. Fine-tuning on this dataset enables the models to better understand user intent, adhere to instructions, and reflect Norwegian cultural nuances.

To ensure high-quality Norwegian instructions, annotators were required to be proficient in Norwegian-specific expressions and knowledgeable about local culture. Given the diversity of the collected data, we recruited annotators with varied backgrounds. A preliminary screening questionnaire was conducted to assess suitability, resulting in 53 NTNU students from different professional backgrounds participating in annotation. Additionally, two students were assigned to quality control. The annotation task lasted three weeks, with annotators submitting their work weekly. Quality control reviewers collected and organized the submissions, returning annotations that failed to meet the standards for revision.

\begin{table}[h!]
\centering
\begin{tabular}{ |p{3.2cm}|p{3cm}|p{3cm}|  }
 \hline
 \multicolumn{3}{|c|}{Task Type (Oppgavetype)}\\
 \hline
 \hline
 Question Answering (Tekstsvar)& Multiple Choice (Flervalg)&Summarization (Oppsummering)\\
 \hline
 4623   & 220    &101\\
 \hline
\end{tabular}
\end{table}

\begin{table}[h!]
\centering
\begin{tabular}{ |p{3cm}|p{3cm}|p{3cm}|p{3cm}| }
 \hline
 \multicolumn{4}{|c|}{Category (Kategori)} \\
 \hline
 \hline
 Norwegian Culture (Norsk kultur)& Expressions (Ord og uttrykk)&Reading Comprehension (Leseforståelse) & Other\\
 \hline
 3611   & 974    &351 &8\\
 \hline
\end{tabular}
\end{table}

We adopt two prompt templates for fine-tuning:
\begin{itemize}
\item With input data: \verb|{instruction}\n\n{inst_input}\nAnswer:|
\item Without input data: \verb|{instruction}\n\nAnswer:|
\end{itemize}

\noindent\textbf{Parameter-Efficient Fine-Tuning with LoRA} To enhance efficiency and scalability across diverse applications, we employed LoRA (Low-Rank Adaptation), a parameter-efficient fine-tuning technique that reduces computational cost while maintaining high performance. Instead of updating all model parameters during fine-tuning, LoRA introduces a pair of small, trainable low-rank matrices within each transformer layer, typically in the attention and feed-forward projection weights. During training, only these low-rank parameters are optimized, while the original pre-trained weights remain frozen. This approach drastically reduces the number of trainable parameters and allows efficient adaptation to new tasks without degrading the base model's general capabilities. 

In our Norwegian LLM pipeline, LoRA was applied to both domain-specific datasets and core language datasets (e.g., English-translated Norwegian QA, summarization, natural language understanding, and dialogue datasets). The resulting adapters\footnote{\url{https://huggingface.co/NorGLM/collections}} capture domain- or task-specific knowledge while preserving the general linguistic competence of the base model. These adapters are modular and can be independently trained, combined, or swapped depending on the application context, enabling flexible deployment and continual expansion of model capabilities.

Together, these instruction-tuning strategies produce Norwegian LLMs that not only follow user instructions more reliably but also demonstrate robust, context-aware behavior across a wide range of domains, making them suitable for both research and practical applications in Norway.

\subsubsection{Preference Optimization}
In addition to instruction-tuning, we leverage Reinforcement Learning with Human Feedback (RLHF) \cite{ouyang2022training} to guide news summarization toward outputs that better reflect human-preferred style on the English-translated Norwegian CNN/DailyMail dataset\footnote{\url{https://huggingface.co/datasets/NorGLM/NO-CNN-DailyMail}}. RLHF is a framework designed to train language models using feedback derived from human judgments, guiding the model toward outputs that are more aligned with human preferences. The standard RLHF workflow consists of three main steps:
\begin{enumerate}[leftmargin=*]
\item \textbf{Supervised Pretraining of the Policy Model:} The base language model is first fine-tuned on task-specific datasets using standard supervised learning to provide a strong initial policy. For news summarization, this involves training the model on available summaries to learn general summarization behavior.

\item \textbf{Reward Model Training:} A reward model is trained to predict human preferences. This typically involves collecting human feedback comparing different outputs for the same input and using it to supervise the reward model. The reward model estimates a scalar reward for any candidate output, representing how closely it aligns with human preferences.

\item \textbf{Policy Optimization via Reinforcement Learning:} The pretrained policy model is updated using reinforcement learning (e.g., Proximal Policy Optimization, PPO), guided by the reward model. The model learns to generate outputs that maximize the predicted human-aligned reward.
\end{enumerate}

\noindent\textbf{Our Approach and Key Difference}
In our adaptation, the main difference lies in step 2, where we train the reward model by estimating semantic similarity between the candidate summary and the human-annotated “golden” summary using NorBERT \cite{kutuzov2021large}. Specifically, candidate summaries with higher cosine similarity to the golden summary receive higher rewards. This approach allows the reward model to capture fine-grained semantic alignment between model outputs and human-preferred summaries, rather than relying solely on explicit human preference comparisons.

By integrating NorBERT-based semantic similarity into the reward model, our RLHF pipeline encourages the generation of summaries that not only preserve key content but also closely match the style and structure preferred by human annotators, improving readability and informativeness in Norwegian news summarization.

\subsection{Deployment Inference} \label{sec:deployment_inference}
This section describes how to deploy and run inference with NorwAI LLMs using the Transformers library, and outlines strategies for efficient quantized deployment as well as hosting options.

\subsubsection{Transformers Example} \label{sec:deployment_inference_hf}
The following Python example demonstrates loading a Norwegian instruction-tuned model and generating responses using Hugging Face Transformers:

\begin{lstlisting}[language=Python, breaklines=true, basicstyle=\ttfamily\footnotesize, backgroundcolor=\color{gray!10}]
from transformers import AutoTokenizer, AutoModelForCausalLM

model_id = "NorwAI/NorwAI-Mistral-7B-instruct"
token = "<HF access token>"

tokenizer = AutoTokenizer.from_pretrained(model_id, token=token)
model = AutoModelForCausalLM.from_pretrained(model_id, token=token, device_map="auto")

prompt = "Hva er Norges lengste elv?\n\nSvar:"
inputs = tokenizer(prompt, return_tensors="pt")
outputs = model.generate(**inputs, max_new_tokens=100)
print(tokenizer.decode(outputs[0], skip_special_tokens=True))
\end{lstlisting}
Notes and best practices: 1) $device\_map=``auto"$ automatically assigns model layers to available devices. 2) Adjust $max\_new\_tokens$ to control output length. 3) For interactive applications, consider streaming generation or using stopping criteria to prevent overly long outputs.

\subsubsection{Quantization}
We support GGUF quantization formats (Q3, Q4, Q5, Q6, Q8) for efficient deployment. Each quantization level provides a trade-off between memory usage and numerical precision:
\begin{itemize}
\item \textbf{Lower Q levels (Q3, Q4):} Reduced memory footprint with slightly lower precision.
\item \textbf{Higher Q levels (Q6, Q8):} Retain higher accuracy while requiring more memory.
\end{itemize}
Users can select the quantization level depending on performance requirements and available resources.

\subsubsection{Alternative Hosting and Deployment Options}
In addition to Transformers-based inference, our models can be hosted using modern LLM serving frameworks and platforms:
\begin{itemize}
\item \textbf{vLLM:} Enables efficient, high-throughput inference with batching and low-latency support.

\item \textbf{Ollama:} Provides simple hosting and API access for LLMs on local or cloud infrastructure.

\item \textbf{Telenor AI Factory:} We are collaborating with Telenor AI Factory to deploy Norwegian LLMs at scale, supporting enterprise-grade applications and secure, managed hosting.
\end{itemize}

\section{Evaluation and Benchmarking}
\subsection{Norwegian Benchmark Dataset - NLEBench}
To systematically evaluate Norwegian LLMs, we developed NLEBench, a comprehensive benchmark dataset covering a wide range of real-world NLP tasks. The datasets are drawn from three main sources: existing Norwegian datasets, machine-translated datasets using the Google Translation API, and manually annotated datasets. Our native Norwegian colleagues evaluated random samples from both Google Translation\footnote{\href{https://cloud.google.com/translate/docs}{https://cloud.google.com/translate/docs}} and an alternative free translation API\footnote{\href{https://pypi.org/project/translators/}{https://pypi.org/project/translators/}}, finding that Google Translation generally performs better, particularly for ambiguous words and long passages. Table \ref{NLEBench-table} summarizes the characteristics, coverage, and statistics of these datasets.
\begin{table*}[t]
\small
  \caption{Overview of the NLEBench dataset.}
  \vspace{0.5\baselineskip}
  \label{NLEBench-table}
  \centering
  \scalebox{0.94}{
  \begin{tabular}{m{3.5cm}m{4.05cm}<{\centering}m{4.55cm}<{\centering}m{2.47cm}<{\centering}}
    \toprule
     Datasets  & Size (\#Samples) & \multicolumn{1}{c}{Task} & Source  \\ \midrule
    \hlineB{1.5}
    \rowcolor{lightgray!50} \multicolumn{4}{c}{Existing Datasets}\\ \hlineB{1}
    NO-Alpaca & 51.942K  & Instruction Finetuning &  \href{https://huggingface.co/datasets/NorGLM/NO-Alpaca-Plus}{Link}  \\
    NO-BoolQ & 12.697K  & Question Answering & \href{https://huggingface.co/datasets/NorGLM/NO-BoolQ}{Link} \\
    NO-QNLI & 110.206K  & Natural Language Inference & \href{https://huggingface.co/datasets/NorGLM/NO-QNLI}{Link} \\
    NO-MRPC & 4076  & Paraphrase & \href{https://huggingface.co/datasets/NorGLM/NO-MRPC}{Link} \\
     \hlineB{1}
    \rowcolor{lightgray!50} \multicolumn{4}{c}{Automatic Machine Translated Datasets (Ours)}\\ \hlineB{1}
    NO-ConvAI2 & 19.845K & Open-domain Conversation & \href{https://huggingface.co/datasets/NorGLM/NO-ConvAI2}{Link}   \\
    NO-CNN/DailyMail & 76.468K & Summarization & \href{https://huggingface.co/datasets/NorGLM/NO-CNN-DailyMail}{Link}  \\
    \multirow{2}{*}{NO-CrowS-Pairs} & 1677 & Bias Detection &  \href{https://huggingface.co/datasets/NorGLM/NO-CrowS-Pairs}{Link}   \\
     & 1508 & Toxicity Detection  & \href{https://huggingface.co/datasets/NorGLM/NO-CrowS-Pairs}{Link}   \\
    \hlineB{1}
    \rowcolor{lightgray!50} \multicolumn{4}{c}{Human Annotated Datasets (Ours)}\\ \hlineB{1}
    NO-Alpaca (extra) & 110  & Instruction Finetuning & \href{https://huggingface.co/datasets/NorGLM/NO-Alpaca-Plus}{Link}  \\
    NO-Multi-QA-Sum & 467 Summaries, 2755 Dialogues & Multi-task Learning & \href{https://huggingface.co/datasets/NorGLM/NO-Multi-QA-Sum}{Link} \\
    \bottomrule
  \end{tabular} }
  \vspace{-2pt}
\end{table*}

\begin{table*}[!htbp]
\small
\centering
  \caption{Experimental Results on the Conversation Task.}
  \vspace{0.5\baselineskip}
  \label{tab:conv}
  \setlength\tabcolsep{3.2pt}
  \scalebox{0.94}{
  \begin{tabular}{m{1.85cm}<{\centering}m{2.0cm}<{\centering}m{1.55cm}<{\centering}m{1.78cm}<{\centering}m{2.73cm}<{\centering}m{1.7cm}<{\centering}m{1.85cm}<{\centering}m{1.1cm}<{\centering}}
    \toprule
    Metrics/Models & NorGPT-369M & NorGPT-3B & NorLlama-3B & NorGPT-3B-continue & NorGPT-23B & NB-GPT-J-6B & GPT-3.5\\
    \midrule
    BLEU & 3.37 & 4.14 & 3.82 & 3.63 & \textbf{4.28} & 3.87 & 2.14 \\
    ROUGE-1  & 16.94 & \textbf{17.09} & 15.20 & 16.47 & 16.72 & 17.05 & 10.82\\
    ROUGE-L & 16.21 & \textbf{16.33} & 14.53 & 15.73 & 15.95 & 16.26 & 9.96\\
    Dist-4 & \textbf{86.54} & 84.68 & 82.47 & 86.33 & 84.41 & 85.83 & 85.80\\
    MAUVE & 0.56 & \textbf{0.87} & 0.61 & 0.71 & 0.64 & 0.68 & 0.72\\
    \bottomrule
  \end{tabular}}
\vspace{-2pt}
\end{table*}

\begin{table*}[!htbp]
\small
\centering
  \caption{Experimental Results on the News Summarization Task.}
  \vspace{0.5\baselineskip}
  \label{tab:sum}
  \setlength\tabcolsep{3.8pt}
  \scalebox{0.94}{
  \begin{tabular}{m{1.7cm}<{\centering}m{1.6cm}<{\centering}m{1.1cm}<{\centering}m{1.65cm}<{\centering}m{1.6cm}<{\centering}cm{1.7cm}<{\centering}cm{1.1cm}<{\centering}}
    \toprule
    Metrics/Models & NorGPT-369M & NorGPT-3B & NorLlama-3B & NorGPT-3B-continue & NorGPT-23B & NorGPT-3B-RLHF & NB-GPT-J-6B & GPT-3.5\\
    \midrule
    BLEU & 2.38 & 2.61 & 0.68 & 2.72 & 1.90 & \textbf{5.41} & 4.35 & 4.38 \\
    ROUGE-1  & 20.97 & 20.31 & 12.32 & 20.53 & 22.44 & 23.01 & 25.64 & \textbf{26.00}\\
    ROUGE-L & 19.68 & 19.05 & 11.56 & 19.26 & 21.13 & 21.63 & 24.25 & \textbf{24.28}\\
    Dist-4 & 95.32 & 94.43 & 92.62 & 94.35 & \textbf{97.66} & 92.18 & 96.41 & 97.13\\
    MAUVE & 0.57 & 0.62 & 0.75 & 0.64 & 0.50 & \textbf{21.03} & 0.65 & 4.38\\ 
    \bottomrule
  \end{tabular}}
\vspace{-2pt}
\end{table*}

\begin{table*}[!htbp]
\small
\centering
  \caption{Experimental Results on the NLU Tasks.}
  \vspace{0.5\baselineskip}
  \label{tab:nlu}
  \setlength\tabcolsep{3.7pt}
  \scalebox{0.94}{
  \begin{tabular}{cccccccc}
    \toprule
    Datasets & Metrics & NorGPT-369M & NorGPT-3B & NorLlama-3B & NorGPT-3B-continue & NorGPT-23B & NB-GPT-J-6B\\
    \midrule
    \multirow{2}{*}{NO-BoolQ} & Accuracy & 58.6 & 60.6 & 56.2 & 58.5 & \textbf{63.2} & 56.7 \\
    & F1 score & 47.8 & 50.3 & 49.0 & 46.7 & \textbf{52.5} & 52.5\\ \hline
    \multirow{2}{*}{NO-QNLI} & Accuracy & 75.8 & 76.4 & 61.7 & 76.9 & 79.7 & \textbf{84.1}\\
    & F1 score & 75.7 & 76.3 & 61.7 & 76.8 & 79.7 & \textbf{84.1}\\ \hline
    \multirow{2}{*}{NO-MRPC} & Accuracy & 71.0 & 68.8 & 66.8 & 69.5 & \textbf{73.7} & 71.7\\
    & F1 score & 54.5 & 46.1 & 52.0 & 55.1 & 64.4 & \textbf{66.6}\\
    \bottomrule
  \end{tabular}}
\vspace{-2pt}
\end{table*}

\begin{table*}[!htbp]
\small
\centering
  \caption{Experimental Results on the NorEval Tasks.}
  \vspace{0.5\baselineskip}
  \label{tab:NorEval}
  \setlength\tabcolsep{3.5pt}
  \scalebox{0.94}{
  \begin{tabular}{m{3.2cm}m{1.45cm}<{\centering}m{1.45cm}<{\centering}m{1.6cm}<{\centering}m{1.5cm}<{\centering}m{1.0cm}<{\centering}m{1.45cm}<{\centering}m{1.6cm}<{\centering}m{1.0cm}<{\centering}}
    \toprule
    Tasks/Models & NorwAI-Mistral-7B & NorwAI-Llama2-7B & NorBLOOM-7B-warm & NorMistral-7B-scratch & Viking-7B & NorMistral-7B-warm & NorMistral-7B-warm-IT & Mistral-7B\\
    \midrule
    Overall & \textbf{45.5} & 44.1 & 35.6 &  38.5 & 41.9 & 43.6 & 40.9 & 39.7 \\ 
    Borda’s Count $\uparrow$  & \textbf{69.0} & 59.0 & 28.0 & 32.0 & 47.0 & 61.0 & 13.0 & 38.0\\ \hline
    Norwegian language knowledge & 47.2 & 47.9 &  51.8 &  53.2 & 51.3 & \textbf{59.2} & 16.9 & 23.4\\ 
    Sentiment analysis & 70.7 & 66.3 &  40.8 & 57.5 & 59.5 & 68.7 & 77.2 & \textbf{77.7}\\ 
    Commonsense reasoning & \textbf{35.9} & 29.8 & 23.5 & 27.7 & 27.4 & 34.0 & 35.2 & 21.1\\ 
    Truthfulness & 36.7 & 30.2 &  39.1 &  40.3 &  26.6 & 31.6 & 24.7  & \textbf{46.0}\\ 
    Norwegian-specific \& world knowledge & 39.5 & 35.4 &  23.3 & 25.4 & 25.0 & 38.7 & \textbf{49.3} & 43.5\\ 
    Reading Comprehension & 37.1 & 38.8 & 23.9 & 22.3 &  25.9 & 40.7 & 23.4 & \textbf{47.1}\\ 
    Text summarization &  31.9 & 37.5 & 35.6 & 35.9 &  49.4 & 33.0 & \textbf{54.8} & 29.5\\
    Instruction following &  37.7 & 37.7 &  13.9 & 14.9 & 38.7 & 14.6 & \textbf{56.1} & 11.6\\ 
    Machine translation &  \textbf{73.2} & 72.9 & 68.8 &  69.7 & 73.0 & 72.0 & 30.5 & 57.5\\
    \bottomrule
  \end{tabular}}
\vspace{4pt}
\end{table*}

\subsection{Quantitative analysis}
We conducted a comprehensive quantitative evaluation of our Norwegian LLMs using NLEBench and other recent Norwegian benchmarking datasets, such as NorEval \cite{mikhailov2025noreval}. These benchmarks provide a diverse set of tasks and metrics, enabling a robust assessment of model performance across multiple linguistic and reasoning dimensions.

\subsubsection{Evaluation Results on Conversation Task}
As shown in Table \ref{tab:conv}, all models, except for GPT-3.5-Turbo, perform quite similarly. Notably, the NorGPT-3B model achieves the best results across multiple evaluation metrics, while the NorGPT-23B model only shows an advantage in BLEU scores. GPT-3.5-Turbo, although specifically curated for conversational purposes, did not exhibit the advantages expected from its extensive knowledge base. This may be because the knowledge of other languages in GPT-3.5-Turbo cannot be directly transferred to understanding Norwegian conversations, highlighting the unique linguistic properties of the Norwegian language.

\subsubsection{Evaluation Results on News Summarization}
In Table \ref{tab:sum}, GPT-3.5-Turbo and NB-GPT-J-6B outperform our NorGPTs on BLEU and ROUGE metrics. This suggests a substantial number of expression patterns resembling news articles in their pre-training datasets. This is plausible given that their datasets likely include a diverse range of newspapers, magazines, and government reports. Additionally, this trend is evident in common test samples, where GPT-3.5-Turbo tends to generate more formal language compared to conversational language. Despite this, we observed that the models' performance improves after reinforcement learning, especially in replicating the word distribution of human writing and generating summaries of similar length. This is supported by the highest scores in MAUVE and BLEU. Although the model with reinforcement learning may not always surpass the fine-tuned model in accuracy, it actively strives to mimic human writing patterns.

\subsubsection{Evaluation Results on NLU tasks}
Table \ref{tab:nlu} reports the results on NLU tasks. Among NorGLMs, NorGPT-23B model consistently outperforms others on different NLU datasets across both evaluation metrics. However, NB-GPT-J-6B performs better on the NO-QNLI benchmark and achieves a higher F1-score on the NO-MRPC benchmark.

\subsubsection{Evaluation Results on NorEval}
Table \ref{tab:NorEval} shows the Borda count and normalized performance scores of the Norwegian LLMs with 7 billion parameters across all task categories in NorEval. Note that these results are reported in the NorEval paper \cite{mikhailov2025noreval}. Among the evaluated models, NorMistral-7B-scratch was trained entirely from scratch on Norwegian text, while NorBLOOM-7B-warm, NorMistral-7B-warm, and NorMistral-7B-warm-IT were initialized from existing pretrained models (warm starts) and further adapted to Norwegian data. The \emph{-IT} suffix denotes additional instruction tuning following the warm-start adaptation phase. We find that NorwAI-Mistral-7B achieves the best overall performance across most task categories among the 7 billion parameter models. Of all the models tested, the 11 billion parameter model NorMistral-11B achieved the best result.

\subsubsection{Evaluation Results on Schibsted News summarization}
A human evaluation was conducted by Schibsted on 51 news articles, each accompanied by three summaries: one written by a professional journalist and two machine-generated summaries - one by GPT-4 and one by our Mixture-of-Experts model, NorwAI-Mixtral-8x7B-Instruct. 
\begin{wrapfigure}{r}{0.52\textwidth}
  \begin{center}
    \includegraphics[width=0.478\textwidth]{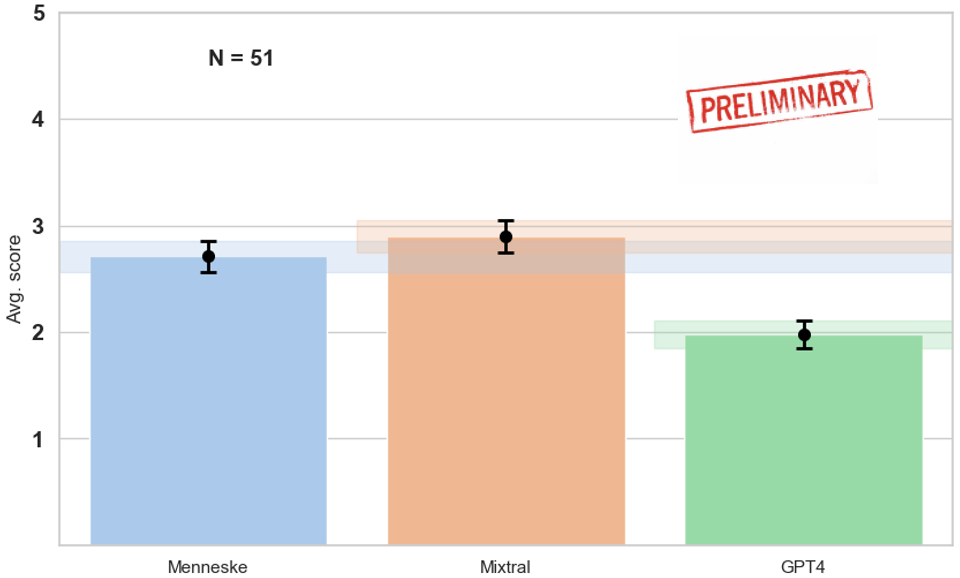}
  \end{center}
  \caption{Human evaluation results on Schibsted news summarization.}
  \label{fig:schibsted_eval}
\end{wrapfigure}
The evaluators were asked to assess the quality of each summary on a five-point Likert scale (1-5), considering factors such as informativeness, coherence, and readability. The evaluation followed a blind review protocol, ensuring that annotators did not know which system produced each summary.

As shown in Figure~\ref{fig:schibsted_eval}, NorwAI-Mixtral-8x7B-Instruct achieved the highest average score, outperforming both GPT-4 and the journalist-written summaries. These preliminary results indicate that the model can produce fluent, coherent, and contextually appropriate Norwegian news summaries that meet professional journalistic standards.

\section{Model Selection and Usage Recommendations}
The NorLLM family comprises models of varying sizes, architectures, and post-training objectives. Selecting the appropriate model depends on task complexity, computational resources, and interaction requirements. The following guidelines summarize recommended usage scenarios.

\subsection{Lightweight and General-Purpose Use}

\textbf{NorGPT-369M} - Recommended for educational use, rapid prototyping, or environments with limited GPU memory (< 8 GB). Suitable for tokenization studies, sentence completion, and basic language modeling.

\textbf{NorGPT-3B / NorLlama-3B} - Balanced between efficiency and accuracy; appropriate for embedded applications, chatbots, and small-scale summarization or classification tasks. Both models are trained from scratch on Norwegian data.

\subsection{Advanced Comprehension and Generation}

\textbf{NorGPT-23B (base)} - Suited for document-level comprehension, long-form text generation, and natural language understanding. Recommended for research or enterprise environments with sufficient compute (> 48 GB GPU memory).

\textbf{NorwAI-Mistral-7B (base)} - A strong dense model for advanced comprehension and fluent Norwegian text generation. It outperforms NorwAI-Llama2-7B on NorEval metrics and serves as a reliable backbone for high-quality analysis and synthesis tasks.

\textbf{NorwAI-Mixtral-8x7B (base)} - Designed for long-context generation (up to 32k tokens) and reasoning-heavy synthesis. The mixture-of-experts architecture dynamically activates two experts per token, improving performance and output quality on diverse downstream tasks relative to the NorwGPT series.

\textbf{NorwAI-Llama2-7B (base)} - Suitable for comprehension-focused or mixed-language generation. Preferred when Llama2 compatibility or cross-lingual robustness is required, though it performs slightly below Mistral and Mixtral in long-form fluency.

\subsection{Instruction-Following and Dialogue Applications}

\textbf{NorwAI-Mistral-7B-Instruct} - Recommended as the default model for most Norwegian interactive and instruction-following tasks. It balances accuracy and latency, supports a 4k-token context, and provides reliable performance across a wide range of practical use cases.

\textbf{NorwAI-Mixtral-8x7B-Instruct} - Ideal for complex assistant-style dialogue, reasoning-intensive tasks, and enterprise virtual-agent deployments.

\textbf{NorwAI-Magistral-24B} - Performaning best on reasoning-heavy applications, combining broad knowledge with
extended context.
Especially designed for multi-step reasoning, and tasks requiring long-context understanding (up to 40k tokens). Recommended for research when longer inputs (16-32k tokens) or more complex reasoning are needed.

\subsection{Domain and Resource Considerations}

\textbf{Low-resource environments:} Quantized GGUF versions (Q4 or Q5) of Mistral-7B-Instruct are recommended for laptop or edge inference.

\textbf{High-throughput serving:} Deploy Mixtral-8x7B or Magistral-24B with vLLM or the Telenor AI Factory for scalable inference.

\textbf{Domain adaptation:} Apply LoRA fine-tuning on top of Mistral-7B-Instruct or Mixtral-8x7B-Instruct for legal, healthcare, or educational applications to reduce compute cost.

\textbf{Minority-language tasks:} Prefer NorwAI-Magistral-24B, trained on NorLLM\_Corpus\_V3, which includes substantial Nynorsk and Sámi data. NorwAI-Mistral-7B and NorwAI-Mixtral-8x7B were trained on NorLLM\_Corpus\_V2, which provides limited coverage but still basic support for mixed Norwegian text.

\section*{Acknowledgments}
This work was funded by SFI NorwAI (Centre for Research-based Innovation, $309834$). The authors gratefully acknowledge the financial support from the Research Council of Norway and the partners of SFI NorwAI. We sincerely thank the IDUN team at NTNU \cite{sjalander2019epic} for providing essential computational resources, and Schibsted, NRK, LTG UIO, and the National Library of Norway (Nasjonalbiblioteket) for supplying the crucial datasets and support for our research.

\appendix
\section{Reading Comprehension Task Formulation} 
This task is designed to assess a model's ability to understand, interpret, and reason about Norwegian texts. The model is presented with a short passage followed by a question that requires comprehension of both explicit and implicit information in the text. The expected response is a concise free-text answer demonstrating the ability to extract relevant details and infer logical conclusions from context.

\begin{itemize}
    \item \textbf{Instruksjon:} ``Ifølge den følgende teksten, hva er viktig å gjøre før man besøker Norge?''
    \item \textbf{Input:} ``Norge kan besøkes til alle årstider, men det er viktig å være klar over at de ulike årstidene gjør noen aktiviteter enklere eller vanskeligere å gjennomføre. Ønsker du å se nordlyset bør du reise i de mørkere og kaldere månedene (oktober - mars), noe som gjør at du kanskje ikke ser så mye sollys og varme. Det vanskelig å få i både pose og sekk, men man kommer langt med åpent sinn!''
    \item \textbf{Output:} ``Bestemme seg for hvilke aktiviteter man vil gjøre i Norge.''
    \item \textbf{Kategori:} ``Leseforståelse''
    \item \textbf{Domene:} ``Annet''
    \item \textbf{Oppgavetype:} ``Tekstsvar''
\end{itemize}

\section{Summarization Task Formulation}
This task evaluates a model's ability to understand, condense, and rephrase informational text in Norwegian. The goal is to produce a concise summary that preserves the key ideas, causal relations, and factual content of the source passage while omitting unnecessary details.

\begin{itemize}
    \item \textbf{Instruksjon:} ``Oppsummer følgende tekst kort.''
    \item \textbf{Input:} ``Havet tar opp mye av varmen på planeten og absorberer samtidig en betydelig del av vår menneskeskapte CO2. Det gjør havet surere og varmere - en katastrofe for korallrev, polare områder og det rike mangfoldet av liv som bor der. I følge FNs klimapanel vil alle verdens tropiske korallrev forsvinne dersom verden når en gjennomsnittlig oppvarming på to grader. Selv om vi klarer å begrense oppvarmingen til 1,5 grader, vil likevel mellom 70 og 90 prosent av verdens tropiske korallrev dø. De fleste skallorganismer, inkludert små krepsdyr som er grunnleggende for den marine matkjeden, er sterkt truet av at havet blir surere. Når vannet blir surt, klarer de ikke å lage skallene sine lenger. Dermed kan de forsvinne helt fra havet. Temperaturøkningen på planeten fører til at havisen ved polene på planeten smelter i rekordfart. Under havisen finnes enorme mengder av små hoppekreps, som danner grunnlaget for alt liv ved iskanten. I Barentshavet finnes for eksempel verdens største torskebestand, enorme sjøfuglkolonier og bestander av forskjellige marine pattedyr. Mindre havis fører til alvorlige konsekvenser for artene som har tilpasset seg et liv på, i og under isen gjennom tusenvis av år.''
    \item \textbf{Output:} ``Menneskeskapt CO2 som tas opp i havet fører til varmere og surere hav, som truer verdens korallrev, skallorganismer havisen og en rekke arter og bestander som lever i, på eller under isen i Barentshavet.''
    \item \textbf{Kategori:} ``Leseforståelse''
    \item \textbf{Domene:} ``Naturfag''
    \item \textbf{Oppgavetype:} ``Oppsummering''
\end{itemize}

\section{Multiple Choice Task Formulation}
This task assesses a model's ability to interpret idiomatic expressions and understand nuanced language use in Norwegian. The model is presented with a question and several answer alternatives, where one or more options may be correct. The goal is to identify which alternatives best satisfy the condition stated in the instruction.

\begin{itemize}
    \item \textbf{Instruksjon:} ``Hvilke av disse følelsene forbindes ikke med `å ha sommerfugler i magen'?''
    \item \textbf{Input:} ``A. Nervøsitet\\ \hspace*{35pt} B. Sorg\\ \hspace*{35pt} C. Spent\\ \hspace*{35pt} D. Sinne\\ \hspace*{35pt} E. Glede''
    \item \textbf{Output:} ``B, D''
    \item \textbf{Kategori:} ``Ord og uttrykk''
    \item \textbf{Domene:} ``Språk''
    \item \textbf{Oppgavetype:} ``Flervalg''
\end{itemize}

\section{Question Answering Task Formulation}
This task evaluates a model's ability to answer open-ended factual or explanatory questions in Norwegian. The model is presented with a natural-language question and is expected to generate a coherent and informative free-text response. The focus of this task is on factual recall, contextual understanding, and knowledge grounding in cultural or domain-specific topics.

\begin{itemize}
    \item \textbf{Instruksjon:} ``Jeg hørte at kollegene mine var med på noe som heter "10 på topp" i, kan du fortelle meg hva det er?''
    \item \textbf{Input:} ``N/A''
    \item \textbf{Output:} ``"Ti på topp" er et turtilbud fra Oslo og Akershus Bedriftsidrettskrets der man kan registrere toppene man er på, for så å være med i trekningen av premier. Bakgrunnen for denne konkurransen er å få flere ut på tur. Man kan delta alene eller i grupper, og man må betale en deltageravgift for å være med. ''
    \item \textbf{Kategori:} ``Norsk kultur''
    \item \textbf{Domene:} ``Sport''
    \item \textbf{Oppgavetype:} ``Tekstsvar''
\end{itemize}


\begin{thebibliography}{9}
\bibitem{brown2020language} Brown, T., Mann, B., Ryder, N., Subbiah, M., Kaplan, J.D., Dhariwal, P., Neelakantan, A., Shyam, P., Sastry, G., Askell, A. and Agarwal, S., 2020. Language models are few-shot learners. Advances in neural information processing systems, 33, pp.1877-1901.

\bibitem{raffel2020exploring} Raffel, C., Shazeer, N., Roberts, A., Lee, K., Narang, S., Matena, M., Zhou, Y., Li, W. and Liu, P.J., 2020. Exploring the limits of transfer learning with a unified text-to-text transformer. Journal of machine learning research, 21(140), pp.1-67.

\bibitem{zhang2024benchmarking} Zhang, T., Ladhak, F., Durmus, E., Liang, P., McKeown, K. and Hashimoto, T.B., 2024. Benchmarking large language models for news summarization. Transactions of the Association for Computational Linguistics, 12, pp.39-57.

\bibitem{zhang2024personalsum} Zhang, L., Liu, P., Henriksboe, M., Lauvrak, E., Gulla, J.A. and Ramampiaro, H., 2024. Personalsum: A user-subjective guided personalized summarization dataset for large language models. Advances in Neural Information Processing Systems, 37, pp.99333-99346.

\bibitem{he2024exploring} He, Z., Liang, T., Jiao, W., Zhang, Z., Yang, Y., Wang, R., Tu, Z., Shi, S. and Wang, X., 2024. Exploring human-like translation strategy with large language models. Transactions of the Association for Computational Linguistics, 12, pp.229-246.

\bibitem{guo2025deepseek} Guo, D., Yang, D., Zhang, H., Song, J., Wang, P., Zhu, Q., Xu, R., Zhang, R., Ma, S., Bi, X. and Zhang, X., 2025. DeepSeek-R1 incentivizes reasoning in LLMs through reinforcement learning. Nature, 645(8081), pp.633-638.

\bibitem{kummervold2021operationalizing} Kummervold, P.E., De La Rosa, J., Wetjen, F. and Brygfjeld, S.A., 2021. Operationalizing a National Digital Library: The Case for a Norwegian Transformer Model. In Proceedings of the 23rd Nordic Conference on Computational Linguistics (NoDaLiDa), pp.20-29.

\bibitem{kutuzov2021large} Kutuzov, A., Barnes, J., Velldal, E., Øvrelid, L. and Oepen, S., 2021. Large-Scale Contextualised Language Modelling for Norwegian. In Proceedings of the 23rd Nordic Conference on Computational Linguistics (NoDaLiDa), pp.30-40.

\bibitem{samuel2023norbench} Samuel, D., Kutuzov, A., Touileb, S., Velldal, E., Øvrelid, L., Rønningstad, E., Sigdel, E. and Palatkina, A., 2023. NorBench-A Benchmark for Norwegian Language Models. In Proceedings of the 24th Nordic Conference on Computational Linguistics (NoDaLiDa), pp.618-633.

\bibitem{de2025impact} De La Rosa, J., Mikhailov, V., Zhang, L., Wetjen, F., Samuel, D., Liu, P., Braaten, R.A., Mæhlum, P., Birkenes, M.B., Kutuzov, A. and Enstad, T., 2025. The Impact of Copyrighted Material on Large Language Models: A Norwegian Perspective. In Proceedings of the Joint 25th Nordic Conference on Computational Linguistics and 11th Baltic Conference on Human Language Technologies (NoDaLiDa/Baltic-HLT 2025), pp.544-560.

\bibitem{liu2024nlebench} Liu, P., Zhang, L., Farup, T., Lauvrak, E.W., Ingvaldsen, J.E., Eide, S., Gulla, J.A. and Yang, Z., 2024. NLEBench+ NorGLM: A Comprehensive Empirical Analysis and Benchmark Dataset for Generative Language Models in Norwegian. In Proceedings of the 2024 Conference on Empirical Methods in Natural Language Processing, pp.5543-5560.

\bibitem{hulora} Hu, E.J., Wallis, P., Allen-Zhu, Z., Li, Y., Wang, S., Wang, L. and Chen, W., 2022. LoRA: Low-Rank Adaptation of Large Language Models. In International Conference on Learning Representations, 2022.

\bibitem{ouyang2022training} Ouyang, L., Wu, J., Jiang, X., Almeida, D., Wainwright, C., Mishkin, P., Zhang, C., Agarwal, S., Slama, K., Ray, A. and Schulman, J., 2022. Training language models to follow instructions with human feedback. Advances in neural information processing systems, 35, pp.27730-27744.

\bibitem{mikhailov2025noreval} Mikhailov, V., Enstad, T., Samuel, D., Farsethås, H.C., Kutuzov, A., Velldal, E. and Øvrelid, L., 2025. NorEval: A Norwegian Language Understanding and Generation Evaluation Benchmark. In Findings of the Association for Computational Linguistics: ACL 2025, pp.3495-3541, Vienna, Austria. Association for Computational Linguistics.

\bibitem{sjalander2019epic} Själander, M., Jahre, M., Tufte, G. and Reissmann, N., 2019. EPIC: An energy-efficient, high-performance GPGPU computing research infrastructure. arXiv preprint arXiv:1912.05848.

\end{thebibliography}
\end{document}